\documentclass[unnumsec,webpdf,contemporary,large]{oup-authoring-template}%

\graphicspath{{Fig/}}

\theoremstyle{thmstyleone}%
\theoremstyle{thmstyletwo}%
\theoremstyle{thmstylethree}%

\usepackage{afterpage} 
\usepackage{booktabs}
\usepackage{multirow}
\usepackage{makecell}  
\usepackage{graphicx}
\usepackage{textpos}
\usepackage{framed}
\usepackage{wrapfig}
\usepackage{multicol}
\usepackage{float}
\usepackage{array}
\usepackage{stfloats}
\usepackage{placeins}
\usepackage{caption}
\usepackage{textgreek}  
\usepackage{tabularx}   
\usepackage{ragged2e}   
\usepackage{threeparttable} 
\usepackage{authblk}  
\usepackage[capitalize]{cleveref} 
\DeclareCaptionLabelFormat{supptabformat}{SuppTable #2} 

\begin{document}

\journaltitle{Journal Title Here}
\DOI{DOI HERE}
\copyrightyear{2022}
\pubyear{2019}
\access{Advance Access Publication Date: Day Month Year}
\appnotes{Paper}

\firstpage{1}


\title[Short Article Title]{A general language model for peptide function identification}

\author{
  Jixiu Zhai$^{1,3,\dagger}$, 
  Zikun Wang$^{1,\dagger}$,
  Chupei Tang$^{1,\dagger}$
  Haitian Zhong$^{4}$, 
  Ziyang Xu$^{5}$, 
  Yuhuan Liu$^{6}$, 
  Shengrui Xu$^{6}$,
  Jingwan Wang$^{2}$, 
  Dan Huang$^{7}$,
  Tianchi Lu$^{1,2\ast}$, 
}

\authormark{Author Name et al.}


\address[1]{School of Mathematics and Statistics, Lanzhou University, 222 South Tianshui Road, Lanzhou 730000, China}

\address[2]{Department of Computer Science, City University of Hong Kong, Kowloon, Hong Kong}

\address[3]{Shanghai Innovation Institute, Shanghai 200231, China}

\address[4]{New Laboratory of Pattern Recognition (NLPR), State Key Laboratory of Multimodal Artificial Intelligence Systems (MAIS), Institute of Automation, Chinese Academy of Sciences (CASIA)}

\address[5]{Department of Mathematics, The Chinese University of Hong Kong, Hong Kong, China}

\address[6]{Cuiying Honors College, Lanzhou University, 222 South Tianshui Road, Lanzhou 730000, China}

\address[7]{Department of Mathematics, Harbin Engineering University, No. 145 Nantong Street, Harbin 150001, China}

\corresp[$\ast$]{Corresponding author. Tianchi Lu, Department of Computer Science, City University of Hong Kong, Kowloon, Hong Kong; School of Mathematics and Statistics, Lanzhou University, 222 South Tianshui Road, Lanzhou 730000, China. \\ Tel:+86-13239620274, \href{email:email-id.com}{tianchilu4-c@my.cityu.edu.hk}}

\received{Date}{0}{Year}
\revised{Date}{0}{Year}
\accepted{Date}{0}{Year}

\abstract{
\textbf{Motivation:} Accurate and generalizable prediction of protein function from sequence is a fundamental challenge in computational biology. Many existing computational methods, however, are limited in their applicability across the diverse spectrum of functional roles.\\
\textbf{Results:} We present PDeepPP, a unified deep learning framework using a pretrained language model with a parallel transformer-CNN architecture to capture both global and local sequence features. We evaluated PDeepPP on a comprehensive benchmark of 33 tasks, focusing on the identification of bioactive peptides (BPs) and post-translational modification (PTM) sites. The model achieves state-of-the-art performance in 25 of these tasks, demonstrating high accuracy and robustness. The framework effectively handles class imbalance and provides interpretable representations, enabling large-scale analysis to support biomedical research and therapeutic discovery.\\
\textbf{Availability and Implementation:} Code, datasets, and pretrained models are publicly available at https://github.com/fondress/PDeepPP and https://huggingface.co/fondress/PDeppPP.}

\maketitle

\section{Introduction}

Accurately identifying a protein's function from its sequence and structure is a foundational challenge in computational biology. This task aims to unveil the roles of unknown proteins and understand their specific contributions to biological processes. Meanwhile, related research also provides critical information for applications such as drug development and disease mechanism analysis, with computational methods increasingly advancing biomedical research and therapeutic discovery\cite{ApplicationforAIandDL,BiopharmaceuticalFormulationsAcceleratedbyMachineLearning,DLfoeCancer}. For example, in drug development, it is necessary to identify bioactive peptides with specific functions (such as antimicrobial or anticancer peptides)\cite{BP2,BioactivePeptides,food_protein-derivedBPs}; in disease research, it is essential to identify post-translational modifications (PTMs) that affect cellular signaling (such as phosphorylation\cite{phosphorylation} or glycosylation\cite{Glycosylation}), which are closely related to cancer, neurodegenerative diseases, and other conditions\cite{PTMindisease}. Traditional experimental methods (such as mass spectrometry), while accurate, are typically costly, time-consuming, and labor-intensive\cite{challengeofPTM,zhipu,zhipucuowu}, limiting large-scale, high-throughput functional screening. Therefore, developing efficient and accurate computational methods to address these limitations has become an important direction in this field.

To overcome these experimental limitations, computational approaches have emerged, evolving from early machine learning methods to more advanced deep learning models\cite{PTM1,PTM2,PTM3}. While early methods improved efficiency, they were often hampered by tedious feature engineering and limited generalization. The recent advent of deep learning has catalyzed significant innovation in proteomics and bioinformatics\cite{protein——DL}, leading to several notable frameworks. For example, UniDL4BioPep\cite{Unidl4Biopep} was developed to uniformly handle classification tasks for diverse bioactive peptides. It pioneered the use of the pretrained protein language model ESM-2 in this domain, simplifying model design. However, this general approach may lack the specificity required to distinguish between bioactive peptides with vastly different sequence patterns, thus limiting its accuracy on some tasks. MusiteDeep\cite{Musite,musite2,musite3,musite4}, a prominent tool for PTM site prediction, uses Convolutional Neural Networks (CNNs) and Capsule Networks to extract locally conserved sequence patterns from sequence windows. While capable of integrating multiple PTM types, its reliance on local features prevents it from capturing the long-range dependencies or global context crucial for identifying some modification sites. This focus on local information restricts its ability to generalize to more complex PTM patterns. Similarly, PhosF3C\cite{phosf3c} was designed for phosphorylation site prediction, employing a feature fusion strategy that combines a fine-tuned protein language model with a Conformer module. Although it excels on its specific task, its efficacy is diminished when applied to highly imbalanced datasets or other PTM types. Its design does not effectively address data sparsity and class imbalance, leading to suboptimal generalization. Overall, these and other specialized predictors\cite{Ptrainsips,Succinylation,methy} typically feature complex designs, lack unified structures, and exhibit limited generalizability across different biological problems\cite{ProgressforPTMprediction}.

To address these limitations, we propose PDeepPP, a unified framework that leverages the pretrained protein language model ESM-2\cite{esm} to extract context-rich embeddings and bypass manual feature engineering. To capture both local and global sequence information, we employ a parallel architecture combining CNNs\cite{CNN} and Transformers\cite{transformer}. The CNN branch identifies local conserved motifs such as modification site patterns, while the Transformer branch uses self-attention to model long-range dependencies across the sequence. This dual-branch design enables comprehensive feature extraction by processing information in parallel and fusing outputs. To handle the class imbalance prevalent in biological datasets, where negative samples often outnumber positive ones, we adopt a loss function inspired by Transductive Information Maximization (TIM)\cite{TIMloss,loss2}. This approach maximizes mutual information between inputs and labels, preventing bias toward the majority class and improving sensitivity for rare positive samples.

Through this design, PDeepPP demonstrates strong performance across 33 benchmark datasets, achieving state-of-the-art results in 25 tasks. The model attains 0.9726 accuracy for antimicrobial peptide identification and 0.9984 for phosphorylation site prediction, with 99.5\% specificity in glycosylation site prediction and substantial reduction in false negatives for antimalarial peptide identification. These results establish PDeepPP as a unified computational tool with potential to accelerate biomedical research and therapeutic target discovery.
\begin{figure*}[htbp]  
    \centering
    \includegraphics[width=1\textwidth]{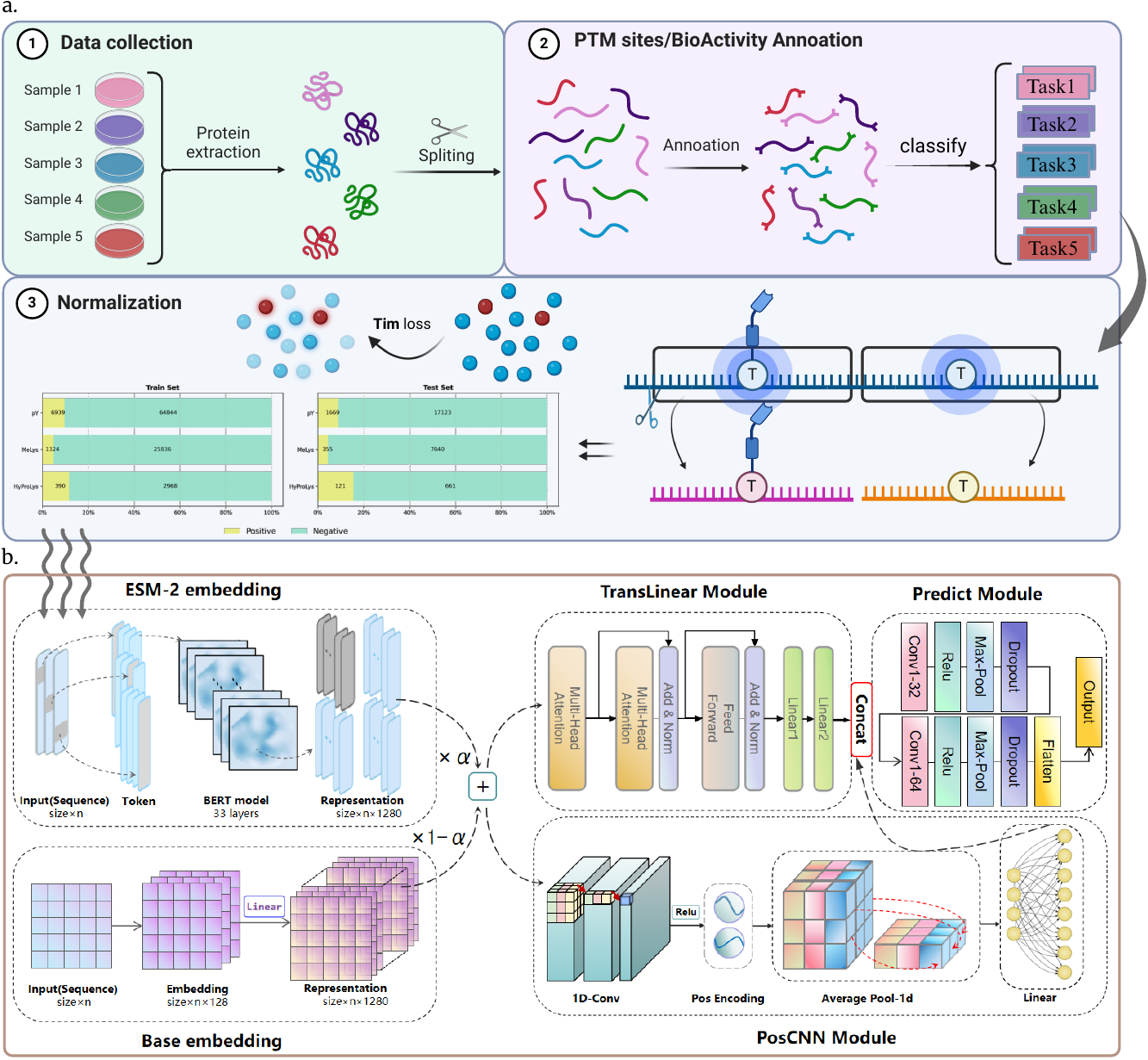}
    \caption{The PDeepPP model usage process consists of three parts: (a) Protein extraction and trimming, peptide chain annotation with task classification, and segment trimming centered on the target amino acid with positive and negative sample classification based on site type. Special loss functions are applied to handle imbalanced datasets. (b) The model framework integrates protein-specific ESM-2 embeddings with a basic tokenizer and weighted linear layers, followed by a parallel network for global and local feature fusion, and convolutional layers for binary classification.}
    \label{fig:model}  
    
\end{figure*}
\begin{table*}[htbp]
\captionsetup{width=0.8\textwidth}
\caption{Benchmark Datasets Sourced from Publications Featuring State-of-the-Art Models}
\label{table:benchmark dataset}
\setlength{\tabcolsep}{10pt}
{\renewcommand{\arraystretch}{1.2}
\begin{tabular}{@{}p{4cm} | p{2cm} p{2cm} | p{2cm} p{2cm} | p{1.8cm}@{}}
\toprule
\textbf{Tasks}& \multicolumn{2}{c|}{\textbf{Training dataset}} & \multicolumn{2}{c|}{\textbf{Test dataset}} & \textbf{Reference} \\
\midrule
 \textbf{Bioactivity} & \textbf{Positives} & \textbf{Negatives} & \textbf{Positives} & \textbf{Negatives} & \\
\midrule
ACE inhibitory activity & 913 & 913 & 386 & 386 & \cite{ACE} \\
DPP IV inhibitory activity & 532 & 532 & 133 & 133 & \cite{DPPIV} \\
Bitter & 256 & 256 & 64 & 64 & \cite{bitter} \\
Umami & 112 & 241 & 28 & 61 & \cite{umami} \\
Antimicrobial activity  & 3876 & 9552 & 2584 & 6369 & \cite{antimicrobial} \\
Antimalarial activity \hfill main & 111 & 1708 & 28 & 427 & \cite{antimalarial} \\
 \hfill alternaive& 111 & 542 & 28 & 135 & \cite{antimalarial} \\
Quorum sensing activity & 200 & 200 & 20 & 20 & \cite{quorum.bibtex} \\
Anticancer activity\hfill main & 689 & 689 & 172 & 172 & \cite{anticancer1,anticancer2} \\
\hfill alternative& 776 & 776 & 194 & 194 & \cite{anticancer1,anticancer2} \\
Anti-MRSA strains activity & 118 & 678 & 30 & 169 & \cite{antimrsa} \\
Tumor T cell antigens & 470 & 318 & 122 & 75 & \cite{TTCA} \\
Blood–Brain Barrier & 100 & 100 & 19 & 19 & \cite{BBP} \\
Antiparasitic activity & 255 & 255 & 46 & 46 & \cite{antiparastic} \\
Neuropeptide & 1940 & 1940 & 485 & 485 & \cite{nuero1,nuero2} \\
Antibacterial activity & 6583 & 6583 & 1695 & 1695 & \cite{Antibacterial} \\
Antifungal activity & 778 & 778 & 215 & 215 & \cite{Antibacterial} \\
Antiviral activity & 2321 & 2321 & 623 & 623 & \cite{Antibacterial} \\
Toxicity & 1642 & 1642 & 290 & 290 & \cite{toxicity} \\
Antioxidant activity & 582 & 541 & 146 & 135 & \cite{antioxidant} \\
\midrule
\textbf{PTMs}\cite{Musite} & \textbf{ Positives} & \textbf{ Negatives} & \textbf{ Positives} & \textbf{ Negatives} & \textbf{Reference} \\
\midrule
Phosphoserine/threonine & 25170 & 473607 & 4847 & 96667 &  \\
Phosphotyrosine & 6939 & 64884 & 1669 & 17123 &  \\
N-linked glycosylation & 52926 & 318092 & 12836 & 76512 &  \\
O-linked glycosylation &567  & 28446 & 143 & 6610 &  \\
N6-acetyllysine & 16222 & 199649 & 3895 & 48653 &  \\
Methylarginine & 3749 & 85090 & 966 & 20825 & \\
Methyllysine & 1324 & 25836 & 335 & 7640 &  \\
S-palmitoylation-cysteine & 2260 & 11904 & 541 & 3172 &  \\
Pyridoxine-carboxylic-acid & 1128 & 7688 & 285 & 1554 &  \\
Ubiquitination & 2528 & 26772 & 581 & 7797 & \\
SUMOylation & 795 & 16209 & 218 & 4036 &  \\
Hydroxylysine & 390 & 2968 & 121 & 661 &  \\
Hydroxyproline & 3931 & 13910 & 892 & 3631 &  \\
\midrule
\textbf{Remaining PTMs} & \textbf{ Positives} & \textbf{ Negatives} & \textbf{ Positives} & \textbf{ Negatives} & \textbf{Reference} \\
\midrule
methylation-G & 627 & 10490 & 165 & 2615 & \cite{methylation-G} \\
methylation-R & 1038 & 1038 & 290 & 290 & \cite{methylation-R} \\
Ubiquitin\_K* & 2528 & 26772 & 581 & 7797 & \cite{ubiquitin} \\
Crotonylation\_K & 6975 & 6975 & 3989 & 3989 & \cite{kcr} \\
\botrule
\end{tabular}
}
\end{table*}

\section{Materials and methods}\label{sec2}

\subsection{Benchmark Datasets and Preprocessing}
\label{subsec1}

\subsubsection*{Data Processing and Task Adaptation}

For each original dataset, we implemented a standard partitioning scheme, allocating 80\% of the data for training and 20\% as a held-out test set. During training, 10\% of the training data was used as a validation set. To address class imbalance present in many datasets, we employed a Transductive Information Maximization (TIM) loss function, which maximizes mutual information between inputs and labels to improve model performance on minority classes.

Our approach was tailored to the prediction task. For Post-Translational Modifications (PTMs), the model focuses on local context around the modification site. For bioactive peptide classification, the entire peptide sequence is treated as a single predictive unit. This task-specific handling ensures appropriate feature extraction for both local (PTM) and global (bioactivity) patterns.

\subsubsection*{Data Sources and Integration Strategy}

We compiled 37 peptide prediction datasets from two review studies and four specialized publications. The datasets are categorized as follows:

\textbf{Bioactivity Datasets (20 sets):} These datasets were compiled from the benchmark established by UniDL4BioPep. They include predictors for angiotensin-converting enzyme (ACE) inhibitory activity (anti-hypertension)\cite{ACE}, dipeptidyl peptidase (DPPIV) inhibitory activity (anti-diabetes)\cite{DPPIV}, bitter\cite{bitter}, umami\cite{umami}, antimicrobial activity\cite{antimicrobial}, antimalarial activity\cite{antimalarial}, quorum-sensing (QS) activity, anticancer activity\cite{anticancer1,anticancer2}, anti-methicillin-resistant S. aureus (MRSA) strains activity\cite{antimrsa}, tumor T cell antigens (TTCA)\cite{TTCA}, blood–brain barrier\cite{BBP}, antiparasitic activity\cite{antiparastic}, neuropeptide\cite{nuero1,nuero2}, antibacterial activity\cite{Antibacterial}, antifungal activity\cite{Antibacterial}, antiviral activity\cite{Antibacterial}, toxicity\cite{toxicity}, and antioxidant\cite{antioxidant} activity.

\textbf{Post-Translational Modification (PTM) Datasets (17 sets):} PTM datasets were curated from benchmarks used by MusiteDeep. All data were originally derived from UniProtKB/Swiss-Prot database\cite{uniprot}. This collection includes datasets for Phosphoserine/threonine, Phosphotyrosine, N-linked glycosylation, O-linked glycosylation, N6-acetyllysine, Methyllysine, S-palmitoylation-cysteine, Pyrrolidone-carboxylic-acid, Ubiquitination, SUMOylation, Hydroxylysine, Hydroxyproline, methylation-G\cite{methylation-G}, methylation-R\cite{methylation-R}, Ubiquitin\_K*\cite{ubiquitin}, and Crotonylation\_K\cite{kcr}.

Detailed information about all benchmark datasets—including the distributions of the training and test splits as well as the specific data sources—is provided in \cref{table:benchmark dataset}. Complete datasets are available on our GitHub repository.

\subsection{Model Architecture}\label{subsec2}

PDeepPP employs a parallel neural network architecture that combines CNNs and Transformers. The TransLinear module uses a Transformer encoder with fully connected layers, while the PosCNN module combines positional encoding with CNN. Outputs from both networks are concatenated and processed through convolutional layers for binary classification. The model integrates ESM-2 protein embeddings with a base tokenizer through weighted linear layers. The complete architecture is shown in \cref{fig:model}.

\subsubsection*{Embedding Strategy}

To integrate pretrained knowledge with task-specific features, we designed a hybrid embedding strategy that dynamically fuses ESM-2 embeddings with a custom BaseEmbedding module.

\textbf{ESM-2 Embeddings:} We employ the ESM-2 model with 650 million parameters as our feature extraction foundation. ESM-2 maps protein sequences into a 1280-dimensional vector space, capturing global long-range dependencies and evolutionary information.

\textbf{BaseEmbedding Module:} To capture task-specific features, we designed the BaseEmbedding module with a task-adaptive architecture. It maps amino acids to a 128-dimensional space, then projects to 1280 dimensions to match ESM-2 dimensions. This module learns local sequence motifs and functional patterns relevant to specific tasks.

These two representations are fused through a weighted sum:

\begin{equation}
    R_{\text{combined}} = \alpha \cdot R_{\text{ESM-2}} + (1-\alpha) \cdot R_{\text{Base}}
\end{equation}

where $R_{\text{combined}}$ is the final hybrid representation, $R_{\text{ESM-2}}$ and $R_{\text{Base}}$ are the embeddings from respective modules, and $\alpha$ (esm\_ratio) is an adjustable hyperparameter. By selecting optimal $\alpha$ values for different tasks (0.9, 0.95, or 1 in our experiments), the model balances general evolutionary knowledge from ESM-2 with task-specific patterns from BaseEmbedding.

\subsubsection*{Parallel Network for Feature Extraction}

\textbf{TransLinear Module:} This module extracts global features from sequences. It processes input through a multi-head self-attention layer with 8 attention heads to compute inter-positional relationships, capturing long-range dependencies. The output is fed into a Transformer encoder with 4 stacked encoder layers, each using 8-head attention. The module incorporates fully connected networks, residual connections, and layer normalization to ensure training stability and information flow.

\textbf{PosCNN Module:} This module extracts local sequence features. Its core is a one-dimensional convolutional layer with kernel size 3 that scans the sequence to identify local patterns such as conserved motifs. A positional encoding layer preserves positional information. An adaptive average pooling layer aggregates local features into a fixed-size vector, which is transformed by a fully connected layer.

The TransLinear and PosCNN modules process sequences in parallel. The TransLinear module captures global dependencies while the PosCNN module extracts local patterns. Feature vectors from each module are concatenated to form a comprehensive representation integrating both global and local information, which is passed to downstream prediction layers for classification.

\subsubsection*{Loss Function}

To optimize model training, we employ a composite loss function inspired by Transductive Information Maximization (TIM)\cite{TIMloss}. Our approach adapts the core idea of TIM loss for fully supervised learning, performing all calculations on the labeled training set.

The loss function augments standard Cross-Entropy (CE) loss with an empirically weighted Mutual Information term to maximize mutual information between input data and labels, enhancing feature representations and model robustness\cite{loss2}.

The final loss function is:
$$
\mathcal{L}(X;Y) := \lambda \cdot \text{CE} - \hat{H}(Y) + \beta \cdot \hat{H}(Y|X)
$$
where:

\textbf{Cross-Entropy Loss (CE):} Standard supervised loss measuring prediction fidelity against ground-truth labels.
$$
\text{CE} := -\frac{1}{|X|}\sum_{i \in X}\sum_{k=1}^{K}y_{ik}\log(p_{ik})
$$

\textbf{Marginal Entropy ($\hat{H}(Y)$):} Computes entropy of the average predicted class distribution. Maximizing this entropy encourages balanced predictions across classes, preventing the model from collapsing to predict only the majority class, particularly important for imbalanced datasets.
$$
\hat{H}(Y) := -\sum_{k=1}^{K}\hat{p}_{k}\log\hat{p}_{k}
$$

\textbf{Conditional Entropy ($\hat{H}(Y|X)$):} Measures uncertainty of predictions for individual samples. Minimizing conditional entropy encourages high-confidence predictions, helping learn clearer classification boundaries.
$$
\hat{H}(Y|X) := -\frac{1}{|X|}\sum_{i \in X}\sum_{k=1}^{K}p_{ik}\log(p_{ik})
$$

Hyperparameters $\beta$ and $\lambda$ balance contributions of each term. $\beta$ is generally set to 1 to align with standard mutual information definition. $\lambda$ balances cross-entropy loss against information-theoretic regularization. Given diversity across peptide bioactivity tasks in dataset size, class imbalance, and complexity, we dynamically tune $\lambda$ for each task (e.g., within [0.9, 1.0]) to find optimal equilibrium between stable supervision and regularization.

\vspace{1em}
\textit{Note: $|X|$ denotes dataset size, $i$ is sample index, $k$ is class index, $p_{ik}$ is predicted probability, and $y_{ik}$ is one-hot encoded ground-truth label. For binary classification, $K=2$.}

\subsection{Model Evaluation}

To evaluate model performance, we adopted standard metrics including Accuracy (ACC), Balanced Accuracy (BACC), Sensitivity (Sn), Specificity (Sp), Matthews Correlation Coefficient (MCC), Area Under the Receiver Operating Characteristic Curve (ROC AUC), and Area Under the Precision-Recall Curve (PR AUC). These metrics are calculated based on True Positives (TP), False Positives (FP), False Negatives (FN), and True Negatives (TN):

\[
\text{ACC} = \frac{TP + TN}{TP + TN + FP + FN}
\]

\[
\text{Sn} = \frac{TP}{TP + FN}
\]

\[
\text{Sp} = \frac{TN}{TN + FP}
\]

\[
\text{BACC} = 0.5 \times \text{Sn} + 0.5 \times \text{Sp}
\]

\[
\text{MCC} = \frac{(TP \times TN) - (FN \times FP)}{\sqrt{(TP + FN) \times (TN + FP) \times (TP + FP) \times (TN + FN)}}
\]

ROC AUC represents binary classification performance by plotting true positive rate (TPR) against false positive rate (FPR) at various thresholds. The AUC score ranges from 0 to 1, where 1 indicates perfect prediction and 0.5 represents random guessing. ROC AUC is calculated using scikit-learn's `roc\_auc\_score` function:

\[
\text{AUC} = \int_0^1 \text{TPR}(\text{FPR}) \, d(\text{FPR})
\]

PR AUC measures the trade-off between precision and recall. Precision is the proportion of true positives among all positive predictions, while recall (sensitivity) is the proportion of true positives among all actual positives. PR AUC is particularly useful for imbalanced datasets where negatives far exceed positives, providing more informative evaluation than ROC curves. PR AUC is computed using scikit-learn's `average\_precision\_score` function:

\[
\text{PR AUC} = \int_0^1 \text{Precision}(\text{Recall}) \, d(\text{Recall})
\]

Higher PR AUC indicates better performance, particularly in imbalanced prediction problems.

\begin{figure*}[htbp]   
    \centering
    \includegraphics[width=1\textwidth]{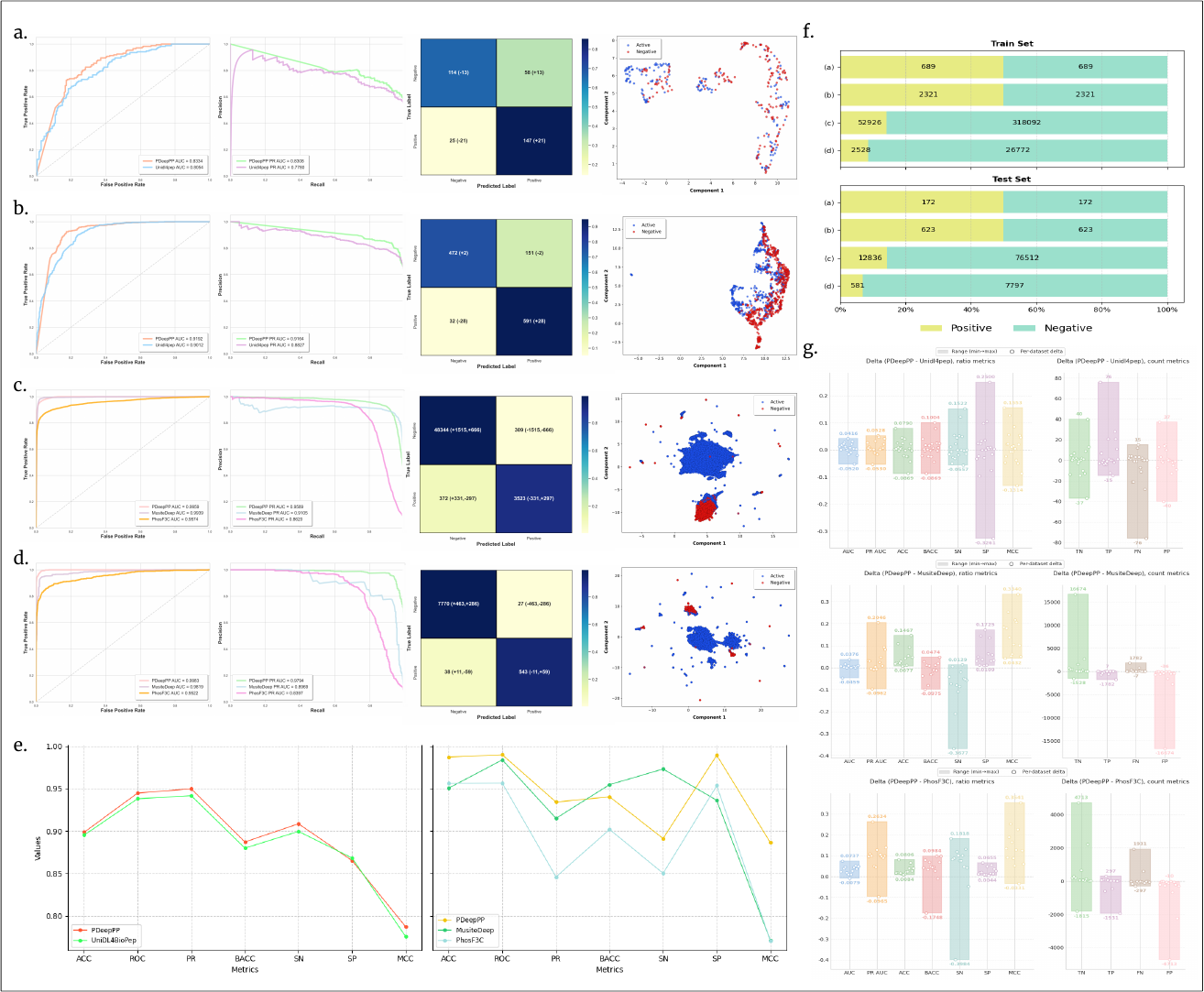}
    \caption{Performance evaluation and analysis of PDeepPP. Panels (a-d) show the performance evaluation on four representative datasets: (a) Anticancer\_main, (b) Antiviral, (c) N6-acetyllysine, and (d) Ubiquitination. For each dataset, the panels from left to right show the ROC curve, PR curve, confusion matrix, and UMAP dimensionality reduction visualization, respectively. Panel (e) is a line chart comparing the average performance metrics of PDeepPP and baseline models across all BP and PTM datasets. Panel (f) illustrates the sample distribution for the training and test sets of the four representative datasets. Panel (g) features incremental bar charts showing the performance delta of PDeepPP against the three baseline models across all evaluation metrics. In the two subplots for each row, each bar chart represents the distribution of numerical differences (PDeepPP minus the baseline model) for that metric across all tasks, including the magnitude and extreme values of all differences. Performance evaluation plots for the remaining tasks are available in the Supplementary Material.}
    \label{fig:contrast}  
\end{figure*}
\afterpage{
\begin{table*}[htbp]
    \addtolength{\oddsidemargin}{-2cm}  
    \addtolength{\evensidemargin}{-1cm} 
    \centering
    \captionsetup{width=0.85\textwidth}
    \caption{Comparison with UniDL4BioPep in bioactive peptides and state-of-the-art models from the same benchmark datasets}
    \label{unidl comparison}
    \setlength{\tabcolsep}{8pt}
    {\renewcommand{\arraystretch}{0.83}
    \begin{tabular}{@{}p{3cm} p{2.5cm} p{1cm} p{1cm} p{1.4cm} p{1cm} p{1cm} p{1cm} p{1cm}@{}}
    \toprule
    \textbf{Bioactivity} & \textbf{Model name} & \textbf{ACC} & \textbf{AUC} & \textbf{PR AUC} & \textbf{BACC} & \textbf{Sn} & \textbf{Sp} & \textbf{MCC}  \\
    \midrule
    \multirow{3}{3cm}{ACE inhibitory activity} 
    & PDeepPP &0.8225 &0.9049 &\textbf{0.9144} &0.8225 &0.8782 &0.7668 &0.6491 \\
    & UniDL4BioPep &0.8433 &0.8918 &0.8728 &\textbf{0.8433} &0.8238 &0.8627 &0.6870  \\
    & mAHTPred & \textbf{0.883} & \textbf{0.951} & & N/A & \textbf{0.894} & \textbf{0.873} & \textbf{0.767}  \\
    
    \midrule
    \multirow{3}{3cm}{DPP IV inhibitory activity}  
    & PDeepPP &\textbf{0.8947} &\textbf{0.9615} &\textbf{0.9654} &\textbf{0.8947} &0.8496 &\textbf{0.9398} &\textbf{0.7927} \\ 
    & UniDL4BioPep &0.8534 &0.9384 &0.9444 &0.8534 &\textbf{0.8647} &0.8421 &0.7069  \\
    & iDPPIV-SCM & 0.797 & 0.847 &  & N/A & 0.789 & 0.805 & 0.594  \\
    
    \midrule
    \multirow{4}{*}{Bitter} 
    & PDeepPP &0.9141 &0.9625 &0.9692 &0.9141 &0.8906 &0.9375 &0.8290 \\
    & UniDL4BioPep &\textbf{0.9375} &\textbf{0.9824} &\textbf{0.9838} &0.9375 &0.9219 &\textbf{0.9531} &0.8754  \\
    & BERT4Bitter & N/A & 0.922  & & \textbf{0.938} & 0.906 & 0.844 & \textbf{0.964} \\
    & iBitter-Fuse & 0.93 & 0.933 &  & N/A & \textbf{0.938} & 0.922 & 0.859  \\
    \midrule
    \multirow{4}{*}{Umami} 
    & PDeepPP &\textbf{0.9213} &\textbf{0.9543} &\textbf{0.9835} &\textbf{0.9330} &0.9016 &\textbf{0.9643} &\textbf{0.8325} \\
    & UniDL4BioPep &0.8764 &0.9417 &0.9760 &0.8326 &\textbf{0.9508} &0.7143 &0.7055  \\
    & UniDL4BioPep-FL & 0.888 & 0.943 &  & 0.883 & 0.892 & 0.875 & 0.733 \\
    & iUmami-SCM & 0.865 & 0.898  & & N/A & 0.714 &  0.934 & 0.679 \\
    \midrule
    \multirow{4}{*}{Antimicrobial activity} 
    & PDeepPP &\textbf{0.9726} &\textbf{0.9947} &\textbf{0.9977} &0.9609 &\textbf{0.9887} &0.9330 &\textbf{0.9329} \\
    & UniDL4BioPep &0.9631 &0.9910 &0.9959 &0.9532 &0.9768 &0.9296 &0.9099 \\ 
    & UniDL4BioPep-FL & 0.96 & 0.991 &  & 0.961 & 0.96 & 0.963 & 0.903 \\
    & TransImbAMP & N/A &  N/A & & \textbf{0.969} & 0.963 & \textbf{0.974} & N/A \\
    \midrule
    \multirow{4}{3cm}{Antimalarial activity (main dataset)} 
    & PDeepPP &0.9758 &\textbf{0.9049} &\textbf{0.9898} &0.8203 &0.9977 &0.6429 &0.7695 \\
    & UniDL4BioPep &\textbf{0.9780} &0.8898 &0.9872 &0.8214 &\textbf{1.0000} &0.6429 &0.7926 \\
    & UniDL4BioPep-FL & \textbf{0.978} &0.898 &  & \textbf{0.965} & 0.979 & 0.95 & \textbf{0.793} \\
    & iAMAP-SCM & \textbf{0.978} & 0.82 &  & 0.826 & 0.654 & \textbf{0.998} &  0.776 \\
    \midrule
    \multirow{4}{3cm}{Antimalarial activity (alternative dataset)} 
    & PDeepPP &0.9877 &\textbf{0.9929} &\textbf{0.9984} &0.9643 &\textbf{1.0000} &0.9286 &0.9566 \\
    & UniDL4BioPep &0.9816 &0.9844 &0.9965 &0.9464 &\textbf{1.0000} &0.8929 &0.9346 \\
    & UniDL4BioPep-FL & \textbf{0.989} & 0.987 &  & \textbf{0.993} & 0.985 & \textbf{1} & \textbf{0.9570} \\
    & iAMAP-SCM & 0.957 & 0.903 &  & 0.896 & 0.808 & 0.985 & 0.834 \\
    \midrule
    \multirow{4}{3cm}{Quorum sensing activity} 
    & PDeepPP &\textbf{0.9750} &\textbf{0.9975} &\textbf{0.9976} &\textbf{0.9750} &\textbf{0.9500} &\textbf{1.0000} &\textbf{0.9512} \\
    & UniDL4BioPep &0.9500 &0.9900 &0.9908 &0.9500 &0.9000 &\textbf{1.0000} &0.9045  \\
    & iQSP & 0.93 & 0.96 &  & N/A & N/A & N/A & 0.86 \\
    & QSPred-FL &  0.943 & 0.945 &  & N/A & 0.935 &  0.95 & 0.885 \\
    \midrule
    \multirow{4}{3cm}{Anticancer activity (main dataset)} 
    & PDeepPP &0.7587 &\textbf{0.8334} &\textbf{0.8308} &0.7587 &\textbf{0.8547} &0.6628 &0.5272 \\
    & UniDL4BioPep &0.7355 &0.8054 &0.7780 &0.7355 &0.7326 &0.7384 &0.4709 \\
    & iACP-FSCM & \textbf{0.825} & 0.812 &  & \textbf{0.825} & 0.726 & \textbf{0.903} & \textbf{0.646} \\
    & AntiCP2.0 & 0.754 & N/A &  & 0.754 & 0.774 & 0.734 & 0.51 \\
    \midrule
    \multirow{4}{3cm}{Anticancer activity (alternative dataset)} 
    & PDeepPP &0.9433 &0.9709 &\textbf{0.9668} &0.9433 &0.9639 &\textbf{0.9227} &0.8874 \\
    & UniDL4BioPep &\textbf{0.9459} &\textbf{0.9711} &0.9630 &\textbf{0.9459} &\textbf{0.9794} &0.9124 &\textbf{0.8938}  \\
    & iACP-FSCM & 0.889 & 0.93  & & N/A & 0.876 & 0.902 & 0.779 \\
    & AntiCP2.0 & 0.92 & N/A &  & N/A & 0.923 & 0.918 & 0.84 \\
    \midrule
    \multirow{4}{3cm}{Anti-MRSA strains activity} 
    & PDeepPP &\textbf{0.9950} &\textbf{0.9998} &\textbf{1.0000} &0.9833 &\textbf{1.0000} &0.9667 &\textbf{0.9803} \\
    & UniDL4BioPep &0.9899 &0.9986 &0.9998 &0.9667 &\textbf{1.0000} &0.9333 &0.9604 \\
    & UniDL4BioPep-FL &0.994 & 0.999 &  & \textbf{0.997} & 0.994 & \textbf{1} & 0.98  \\
    & SCMRSA & 0.96 & 0.986 &  & 0.935 & 0.9 & 0.97 & 0.848 \\
    \midrule
    \multirow{4}{*}{Tumor T cell antigens} 
    & PDeepPP &\textbf{0.7563} &\textbf{0.8198} &\textbf{0.7662} &0.7185 &0.5600 &\textbf{0.8770} &\textbf{0.4680} \\
    & UniDL4BioPep &0.7513 &0.7883 &0.7395 &0.7144 &0.5600 &0.8689 &0.4569  \\
    & UniDL4BioPep-FL & 0.746 & 0.796 &  & \textbf{0.762} & 0.734 & 0.791 & 0.446 \\
    & iTTCA-Hybrid &  0.71 & 0.756  & & N/A & \textbf{0.844} & 0.493 & 0.363 \\
    \midrule
    \multirow{3}{*}{Blood–Brain Barrier} 
    & PDeepPP &\textbf{0.9211} &\textbf{0.9640} &\textbf{0.9680} &\textbf{0.9211} &\textbf{0.9474} &0.8947 &\textbf{0.8433} \\
    & UniDL4BioPep &0.8421 &0.9224 &0.9211 &0.8421 &0.8947 &0.7895 &0.6880  \\
    & BBBpred & 0.7895 & 0.7895  & & N/A & 0.6316 & \textbf{0.9474} & 0.6102  \\
    \midrule
    \multirow{3}{*}{Antiparasitic activity} 
    & PDeepPP &0.7609 &\textbf{0.9267} &\textbf{0.9370} &0.7609 &0.9565 &0.5652 &0.5669 \\
    & UniDL4BioPep &\textbf{0.8478} &0.9045 &0.9258 &\textbf{0.8478} &0.8043 &\textbf{0.8913} &0.6983  \\
    & PredAPP & 0.88 & 0.922 &  & N/A & \textbf{0.978} & 0.783 & \textbf{0.775} \\
    \midrule
    \multirow{4}{*}{Neuropeptide} 
    & PDeepPP &0.9021 &0.9611 &0.9602 &0.9021 &\textbf{0.9010} &0.9031 &0.8041 \\
    & UniDL4BioPep &0.9052 &0.9707 &\textbf{0.9707} &\textbf{0.9052} &0.8784 &0.9320 &0.8115  \\
    & PreNeuroP & 0.897 & 0.954 &  & N/A & 0.886 & 0.907 & 0.794 \\
    & NeuroPred-CLQ & \textbf{0.936} & \textbf{0.988} &  & N/A & 0.897 & \textbf{0.975} & \textbf{0.875}  \\
    \midrule
    \multirow{3}{*}{Antibacterial activity} 
    & PDeepPP &\textbf{0.9478} &\textbf{0.9811} &\textbf{0.9752} &\textbf{0.9478} &0.9617 &0.9339 &\textbf{0.8959} \\
    & UniDL4BioPep &0.9395 &0.9775 &0.9704 &0.9395 &\textbf{0.9687} &0.9103 &0.8806  \\
    & ABPDiscover & 0.935 & 0.975 &  & N/A & 0.912 & \textbf{0.957} & 0.87  \\
    \midrule
    \multirow{3}{*}{Antifungal activity} 
    & PDeepPP &\textbf{0.9535} &\textbf{0.9911} &\textbf{0.9916} &\textbf{0.9535} &0.9442 &\textbf{0.9628} &\textbf{0.9071} \\
    & UniDL4BioPep &0.9512 &0.9862 &0.9838 &0.9512 &0.9628 &0.9395 &0.9026  \\
    & ABPDiscover & 0.942 & 0.988 &  & N/A & 0.921 
    & \textbf{0.963} & 0.884  \\
    
    \midrule
    \multirow{3}{*}{Antiviral activity} 
    & PDeepPP &\textbf{0.8531} &\textbf{0.9192} &\textbf{0.9164} &\textbf{0.8531} &\textbf{0.9486} &0.7576 &\textbf{0.7195} \\
    & UniDL4BioPep &0.8291 &0.9012 &0.8827 &0.8291 &0.9037 &0.7544 &0.6656  \\
    & ABPDiscover & 0.828 & 0.896 &  & N/A & 0.764 & \textbf{0.892} & 0.662  \\
    
    \midrule
    \multirow{3}{*}{Toxicity} 
    & PDeepPP &0.9172 &0.9779 &0.9770 &0.9168 &0.9108 &0.9228 &0.8336 \\
    & UniDL4BioPep &\textbf{0.9603} &\textbf{0.9943} &\textbf{0.9934} &\textbf{0.9608} &\textbf{0.9665} &\textbf{0.9550} &\textbf{0.9205}  \\
    & ATSE & 0.952 & 0.976 &  & N/A & 0.965 & 0.94 & 0.903  \\
    \midrule
    \multirow{3}{*}{Antioxidant activity} 
    & PDeepPP &0.7972 &0.8776 &0.8880 &0.7981 &0.7740 &0.8222 &0.5959 \\
    & UniDL4BioPep &\textbf{0.8221} &\textbf{0.9296} &\textbf{0.9410} &\textbf{0.8229} &\textbf{0.8014} &\textbf{0.8444} &\textbf{0.6454}  \\
    & AntiOxPred-FRS & N/A & 0.79 &  & N/A & N/A & N/A & 0.48 \\
    \bottomrule
    \end{tabular}
    }
\end{table*}

\begin{table*}[htbp]
    \caption{Comparison with the MusiteDeep and PhosF3C in PTMs from the same benchmark datasets}
    \label{table:musite comparison}
    {\renewcommand{\arraystretch}{0.95}
    \begin{tabular}{@{}p{3.5cm} p{2cm} p{1.2cm} p{1.2cm} p{1.5cm} p{1.2cm} p{1.2cm} p{1.2cm} p{1.2cm}@{}}
    \toprule
    \textbf{PTM types} & \textbf{Model} & \textbf{ACC} & \textbf{AUC} & \textbf{PR AUC}  & \textbf{BACC} & \textbf{SN} & \textbf{SP} & \textbf{MCC} \\
    \midrule
    \multirow{3}{*}{Hydroxyproline} & PDeepPP & \textbf{0.9885} & \textbf{0.9992} & \textbf{0.9959} & \textbf{0.9729} & 0.9504 & \textbf{0.9955} & \textbf{0.9557} \\  
     & MusiteDeep & 0.9412 & 0.9937 & 0.9718 & 0.9618 & \textbf{0.9917} & 0.9319 & 0.8187 \\  
     & PhosF3C & 0.9476 & 0.9625 & 0.8892 & 0.8745 & 0.7686 & 0.9803 & 0.7912 \\
    \midrule
    \multirow{3}{*}{Hydroxylysine} & PDeepPP & \textbf{0.9648} & \textbf{0.9915} & \textbf{0.9615} & \textbf{0.9430} & 0.9070 & \textbf{0.9791} & \textbf{0.8887} \\  
     & MusiteDeep & 0.8720 & 0.9866 & 0.9523 & 0.9152 & \textbf{0.9865} & 0.8438 & 0.7084 \\  
     & PhosF3C & 0.8998 & 0.9473 & 0.8725 & 0.8645 & 0.8061 & 0.9229 & 0.6992 \\
     \midrule
    \multirow{3}{*}{Methyllysine} & PDeepPP & \textbf{0.9984} & \textbf{0.9997} &\textbf{0.9972} & \textbf{0.9924} & \textbf{0.9859} & \textbf{0.9990} & \textbf{0.9809} \\  
     & MusiteDeep & 0.9877 & 0.9921 & 0.9649 & 0.9842 & 0.9803 & 0.9881 & 0.8756 \\  
     & PhosF3C & 0.9869 & 0.9768 & 0.9005 & 0.9448 & 0.8986 & 0.9910 & 0.8527 \\
     \midrule
    \multirow{3}{*}{Methylarginine} & PDeepPP & \textbf{0.9918} & \textbf{0.9977} & 0.9565 & 0.9400 & 0.8830 & \textbf{0.9969} & \textbf{0.9016} \\  
     & MusiteDeep & 0.9621 & 0.9949 & \textbf{0.9638} & \textbf{0.9767} & \textbf{0.9928} & 0.9607 & 0.7171 \\  
     & PhosF3C & 0.9794 & 0.9702 & 0.8277 & 0.8969 & 0.8064 & 0.9875 & 0.7665 \\
     \midrule
    \multirow{3}{*}{N-linked glycosylation} & PDeepPP & \textbf{0.9810} & \textbf{0.9970} & \textbf{0.9806} & 0.9644 & 0.9415 & \textbf{0.9873} & \textbf{0.9204} \\  
     & MusiteDeep & 0.9597 & 0.9883 & 0.8874 & \textbf{0.9760} & \textbf{0.9986} & 0.9535 & 0.8579 \\  
     & PhosF3C & 0.9636 & 0.9869 & 0.8812 & 0.9740 & 0.9884 & 0.9596 & 0.8674 \\
     \midrule
    \multirow{3}{*}{N6-acetyllysine} & PDeepPP & \textbf{0.9870} & \textbf{0.9959} & \textbf{0.9589} & 0.9491 & 0.9045 & \textbf{0.9936} & \textbf{0.9049} \\  
     & MusiteDeep & 0.9645 & 0.9939 & 0.9105 & \textbf{0.9760} & \textbf{0.9895} & 0.9625 & 0.8033 \\  
     & PhosF3C & 0.9687 & 0.9574 & 0.8620 & 0.9041 & 0.8282 & 0.9800 & 0.7807 \\
     \midrule
    \multirow{3}{*}{Phosphotyrosine} & PDeepPP & \textbf{0.9944} & \textbf{0.9989} & \textbf{0.9909} & \textbf{0.9818} & 0.9664 & \textbf{0.9971} & \textbf{0.9654} \\  
     & MusiteDeep & 0.9567 & 0.9943 & 0.9417 & 0.9711 & \textbf{0.9886} & 0.9536 & 0.7967 \\  
     & PhosF3C & 0.9486 & 0.9542 & 0.8210 & 0.8964 & 0.8328 & 0.9599 & 0.7194 \\
     \midrule
    \multirow{3}{*}{Phosphoserine/threonine} & PDeepPP & \textbf{0.9647} & 0.9762 & 0.6487 & 0.7774 & 0.5702 & \textbf{0.9845} & 0.5896 \\  
     & MusiteDeep & 0.8180 & 0.9386 & 0.4441 & 0.8749 & 0.9379 & 0.8120 & 0.3836 \\  
     & PhosF3C & 0.9373 & \textbf{0.9841} & \textbf{0.7452} & \textbf{0.9522} & \textbf{0.9686} & 0.9357 & \textbf{0.6227} \\
     \midrule
    \multirow{3}{*}{Pyrrolidone-carboxylic-acid} & PDeepPP & \textbf{0.9804} & 0.9964 & 0.9822 & 0.9626 & 0.9368 & \textbf{0.9884} & \textbf{0.9253} \\  
     & MusiteDeep & 0.9625 & \textbf{0.9978} & \textbf{0.9910} & \textbf{0.9735} & \textbf{0.9895} & 0.9575 & 0.8749 \\  
     & PhosF3C & 0.9641 & 0.9772 & 0.9317 & 0.9358 & 0.8947 & 0.9768 & 0.8642 \\
     \midrule
    \multirow{3}{*}{O-linked glycosylation} & PDeepPP & \textbf{0.9947} & 0.9457 & 0.8231 & 0.8878 & 0.7762 & \textbf{0.9994} & \textbf{0.8631} \\  
     & MusiteDeep & 0.9508 & \textbf{0.9916} & \textbf{0.9193} & \textbf{0.9680} & \textbf{0.9860} & 0.9501 & 0.5291 \\  
     & PhosF3C & 0.9730 & 0.8720 & 0.5597 & 0.8186 & 0.6573 & 0.9799 & 0.5090 \\
     \midrule
    \multirow{3}{*}{S-palmitoylation-cysteine} & PDeepPP & \textbf{0.9908} & \textbf{0.9971} & \textbf{0.9906} & \textbf{0.9885} & \textbf{0.9852} & \textbf{0.9918} & \textbf{0.9639} \\  
     & MusiteDeep & 0.9793 & 0.9965 & 0.9861 & 0.9764 & 0.9723 & 0.9805 & 0.9207 \\  
     & PhosF3C & 0.9666 & 0.9871 & 0.9453 & 0.9544 & 0.9372 & 0.9716 & 0.8728 \\
     \midrule
    \multirow{3}{*}{SUMOylation} & PDeepPP & \textbf{0.9915} & 0.9982 & 0.9713 & 0.9521 & 0.9083 & \textbf{0.9960} & \textbf{0.9123} \\  
     & MusiteDeep & 0.9838 & \textbf{0.9987} & \textbf{0.9755} & \textbf{0.9828} & \textbf{0.9817} & 0.9839 & 0.8600 \\  
     & PhosF3C & 0.9831 & 0.9459 & 0.8725 & 0.9086 & 0.8257 & 0.9916 & 0.8245 \\
     \midrule
    \multirow{3}{*}{Ubiquitination} & PDeepPP & \textbf{0.9922} & \textbf{0.9983} & \textbf{0.9794} & \textbf{0.9656} & 0.9346 & \textbf{0.9965} & \textbf{0.9394} \\  
     & MusiteDeep & 0.9383 & 0.9819 & 0.8969 & 0.9453 & \textbf{0.9535} & 0.9372 & 0.6851 \\  
     & PhosF3C & 0.9511 & 0.9522 & 0.8397 & 0.8965 & 0.8330 & 0.9599 & 0.6866 \\
    \botrule
    \end{tabular}
    }
\end{table*}
}
\section{Results}\label{sec3}
\subsection{Performance evaluation of PDeepPP with notable framewoeks on the independent test datasets of BPs and PTMs}    

To evaluate and validate the robustness and accuracy of PDeepPP, we compared its performance against UniDL4BioPep, MusiteDeep, and PhosF3C on both Bioactive Peptide and Post-Translational Modification prediction tasks, as detailed in \cref{unidl comparison,table:musite comparison}. \cref{fig:contrast}e provides a visual summary of the average performance metrics for each model across both task categories. In tasks where both ACC and AUC were surpassed, for Bioactive Peptide (BP) tasks, PDeepPP demonstrated an average improvement of 2.28\% in Accuracy (ACC), 1.54\% in Area Under the Curve (AUC), and 1.77\% in Precision-Recall Area Under the Curve (PR AUC). For Post-Translational Modification (PTM) tasks, the average improvements were 3.64\%, 0.59\%, and 3.73\% compared to MusiteDeep, and 3.07\%, 3.12\%, and 10.8\% compared to PhosF3C, respectively. Across a total of 33 datasets, our model achieved the highest accuracy on 25 of them.

\cref{fig:contrast}g further quantifies the performance discrepancies between PDeepPP and the baseline models using incremental bar plots. Taking the comparison with PhosF3C on PTM tasks as an example, PDeepPP outperformed in both ACC and SP metrics. Furthermore, the MCC score improved on 12 out of 13 datasets, with 8 of these showing an improvement margin greater than 0.1. Concurrently, count-based metrics indicated a reduction in false positive predictions by PDeepPP across all datasets. On the BP tasks, although the overall performance gap against UniDL4BioPep was not substantial, PDeepPP achieved superior performance in ACC, AUC, and PR AUC on over 60\% of the tasks.

To conduct an in-depth investigation of the model's performance under various scenarios, we selected four representative datasets for analysis: N6-acetyllysine, Antiviral, Ubiquitination, and Anticancer\_main. These datasets encompass a range of typical challenges, from balanced distributions to extreme class imbalance, and from clear biological signals to high heterogeneity. Detailed results are presented in \cref{fig:contrast}, which includes AUC and AUPRC curves, confusion matrices, UMAP visualizations (\cref{fig:contrast}a-d), and sample distribution plots (\cref{fig:contrast}f) for each dataset.

The confusion matrices validate the model's capability in handling class imbalance, as demonstrated in the N6-acetyllysine task (\cref{fig:contrast}c), where negative samples outnumber positive samples by approximately sixfold. On this task, MusiteDeep and PhosF3C generated 1,824 and 976 false positive predictions, respectively. In contrast, PDeepPP reduced the number of false positives to 309 while maintaining a high recall rate. Conversely, on the data-balanced Antiviral task (\cref{fig:contrast}b), the model correctly identified 94.86\% of positive samples and 75.76\% of negative samples, achieving an MCC of 0.7195 (compared to a baseline of 0.6656).

The intrinsic biological characteristics of the data remain a key determinant of the upper limit of model performance. For instance, the Ubiquitination dataset presents a scenario with a clear classification boundary (\cref{fig:contrast}d). Despite class imbalance, the presence of conserved sequence motifs leads to distinct cluster separation in the UMAP visualization, and the model achieves a high MCC of 0.9394. In contrast, the highly heterogeneous Anticancer\_main dataset (\cref{fig:contrast}a), despite being class-balanced, features diverse sequence patterns and a low signal-to-noise ratio. This results in significant cluster overlap in the UMAP visualization and an MCC of only 0.5272. This suggests that the model can learn a stable decision boundary when the data signal is clear, whereas its performance ceiling is constrained in scenarios with higher heterogeneity.

\subsection{Model Interpretability Analysis Reveals Key Sequence Motifs}

\begin{figure*}
    \includegraphics[width=1\textwidth]{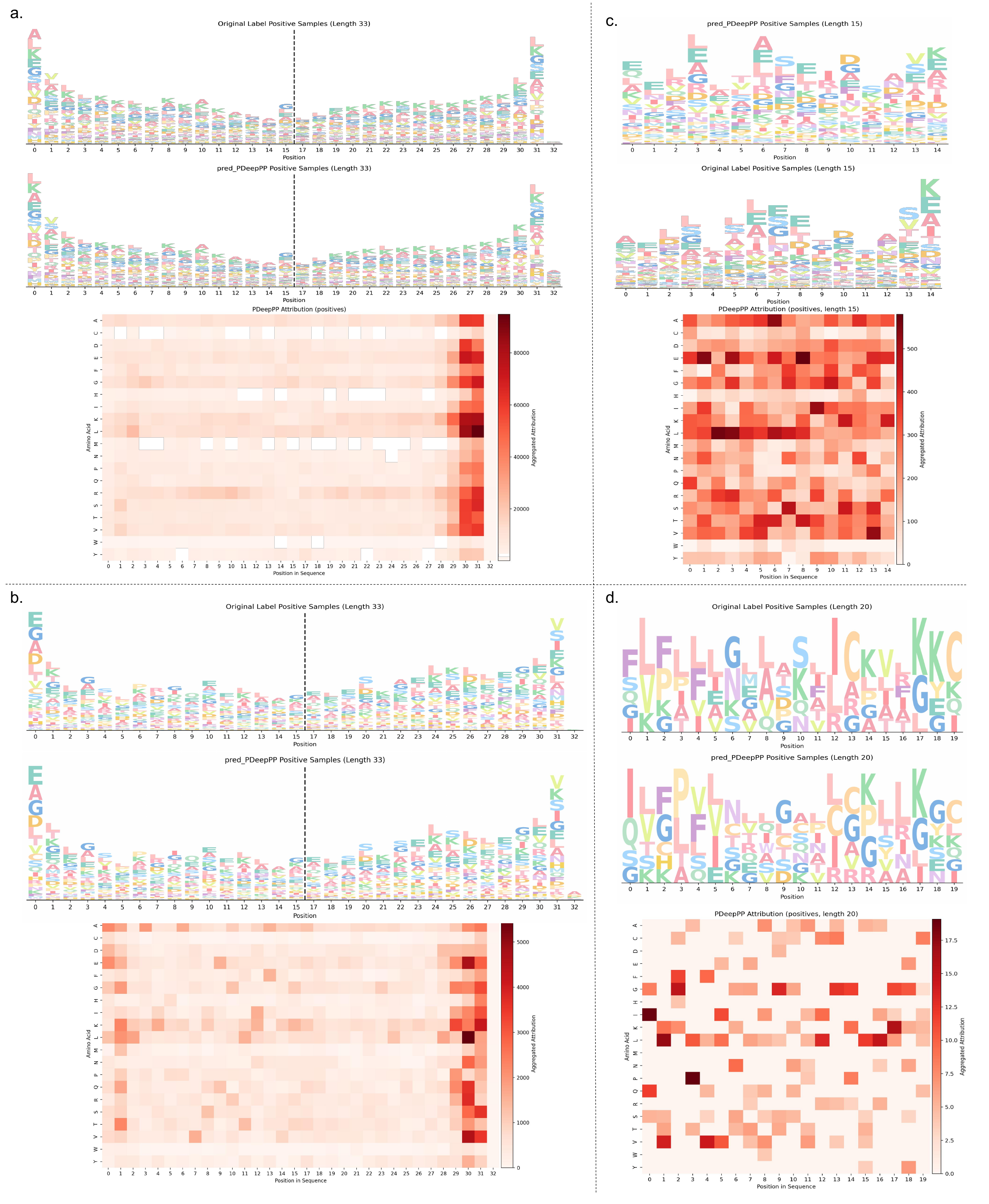}
    \caption{Interpretability analysis of PDeepPP across four datasets. (a) N6-acetyllysine, (b) Ubiquitination, (c) Antiviral, (d) Anticancer\_main. Each panel contains three visualizations from top to bottom: (1) Sequence logo of original positive samples showing amino acid conservation patterns, where letter height represents information content (bits) with highly conserved residues (high information) stacked on top; (2) Sequence logo of PDeepPP-predicted positive samples; (3) Attribution heatmap showing aggregated Integrated Gradients attributions for positive predictions, where color intensity (white to dark red) indicates the importance of each amino acid at each position for model decisions. For the two PTM tasks, Position 16 (central site) is excluded from visualization and marked with a dashed line. For the two BP tasks, due to inconsistent sequence lengths across samples, three representative lengths were selected for each task. The visualizations shown here correspond to the most frequent length. Length distributions for both datasets and interpretability visualizations for the other representative lengths are provided in the Supplementary Material.}
    \label{fig:interpretability}
\end{figure*}

To investigate the sequence patterns learned by PDeepPP, we conducted an interpretability analysis on four representative datasets using a combination of sequence logos and attribution heatmaps, as shown in \cref{fig:interpretability}. Sequence logos show the amino acid conservation at each position for both ground-truth and model-predicted positive samples, while attribution heatmaps quantify the contribution of each amino acid at each position to the final positive prediction. Through comparative analysis, we can assess whether the model successfully captures key data features and the degree of correspondence between the predicted and ground-truth logos.

For PTM tasks, the model successfully learns key sequence features, with the predicted logo showing high consistency with the true distribution. In the N6-acetyllysine task, the model captures key features at both ends of the sequence (upstream positions 0-4 and downstream positions 28-32 show higher contribution scores). The signals learned from these distal regions may be related to maintaining a specific spatial conformation conducive to enzyme recognition and binding. At the same time, the predicted sequence logo is highly consistent with the ground-truth logo, nearly reproducing the distribution pattern of hydrophobic amino acids (A, V, L) and polar neutral amino acids (G, S). In the Ubiquitination task, the model also accurately captures key features at the sequence ends. Its predicted logo successfully replicates the conserved distribution of small molecular weight amino acids (such as A and G) at multiple positions, as seen in the ground-truth logo, indicating that it has learned the overall pattern dominated by these residues.
For BP tasks, the model performs excellently on some conserved sites but shows deviations on more heterogeneous data. In the Antiviral task, the model precisely captures the strong preference for the positively charged Lysine (K) at position 14—the predicted logo matches the ground-truth logo at this site, and the attribution heatmap also verifies the high contribution of this key feature. The model also successfully identifies the conserved pattern of hydrophobic Leucine (L) in the 2-8 region, with the prediction being consistent with the true distribution, and the attribution heatmap showing a continuous dark band for L's contribution in positions 2-8. However, the model's predicted logo fails to reproduce the preference for Arginine (R) at position 9 found in the ground-truth samples. For the Anticancer\_main task, although the model accurately identifies the high conservation of Lysine (K) at position 17 (predicted logo matches the ground truth), a significant deviation occurs at the initial position 0—the ground-truth logo shows a preference for Phenylalanine (F), whereas the model's prediction is concentrated on Isoleucine (I). This may impact prediction accuracy (ACC: 0.7587). Although the ground-truth sequence logo for the Anticancer\_main task appears highly conserved at some positions, the model's predicted MCC (0.5272) is significantly lower than that for the Antiviral task (0.7195). The diversity of anticancer peptide mechanisms and the limited training sample size are likely the main reasons the model struggles to fully capture the true distribution pattern.

\subsection{Systematic Component Analysis Validates the Model Design} 
\subsubsection{Experimental Setup}
To systematically validate the efficacy of each key component within our proposed PDeepPP framework, we designed and conducted a comprehensive series of ablation studies. By systematically removing or replacing the model's key modules, we created six ablated variants to quantitatively assess the contribution of each part to the final performance. These six variants include:
\begin{itemize}
    \item \textbf{w/o embedding}: The BaseEmbedding module was removed($\alpha = 1$).
    \item \textbf{w/o Translinear} and \textbf{w/o PosCNN}: The global dependency capture branch (Translinear) and the local motif extraction branch (PosCNN) were removed, respectively.
    \item \textbf{w/o attention} and \textbf{w/o PosEncoding}: The pre-encoder attention mechanism and the positional encoding module were removed, respectively.
    \item \textbf{w/o loss}: The TIM loss function was replaced with a standard cross-entropy loss function.
\end{itemize}

To ensure a fair baseline in these comparative experiments, the full PDeepPP model used as a reference employed fixed hyperparameters. As hyperparameter optimization statistics showed that the optimal ESM-2 weight $\alpha$ values (0.9, 0.95, and 1.0) occurred with identical frequency across all tasks, the boundary value of 0.9 was selected for this weight in the experiment. Concurrently, the cross-entropy weight $\lambda$ in the TIM loss function was set to 0.95, as it was one of the most frequent central values across all tasks and thus highly representative.

\begin{figure*}[htbp]
        \centering
        \hspace*{-0.05\textwidth}
        \includegraphics[width=1.1\textwidth]{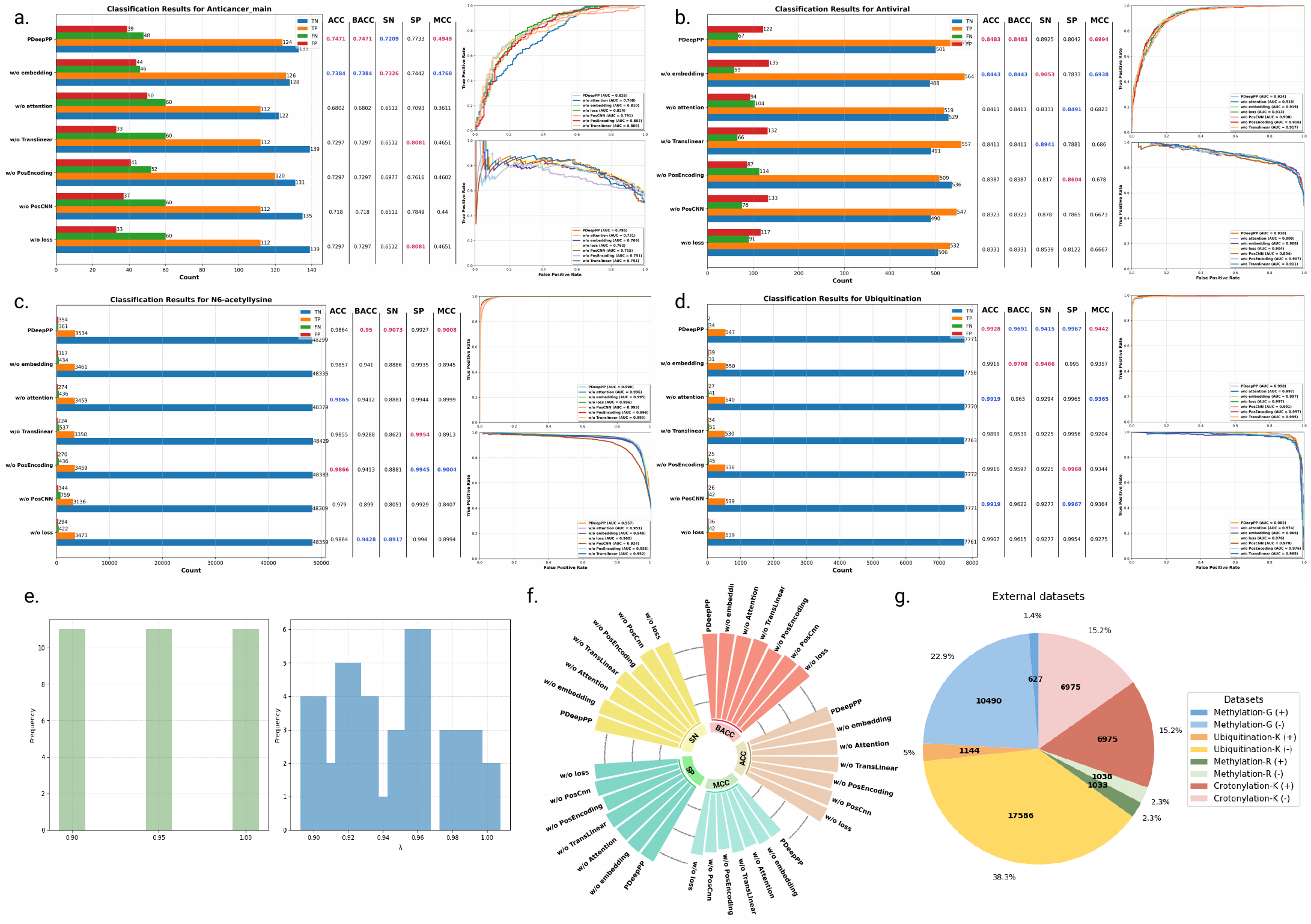} 
        \caption{These sections show the identification results of the PDeepPP model and its six ablated variants on four tasks: Anticancer\_main, Antiviral, N6-acetyllysine, and Ubiquitination. (a) - (d) In each subplot, the bar chart on the left displays the counts for the four confusion matrix metrics (TP, TN, FP, FN); the table in the middle compares the numerical values of different models on various performance metrics (acc, bacc, sn, sp, mcc), where red indicates the best performance and blue indicates the second-best. The right side shows the ROC and PR curves for the models.(e) This section shows the count distribution of the optimal parameter values obtained after training the embedding and loss components on all 33 internal datasets.(f) This section shows the average acc, bacc, sn, sp, and mcc of all models across all datasets.(g) This section shows the counts and proportions of positive and negative samples in the external datasets.}
        \label{fig:ablation_combine} 
\end{figure*}
\subsubsection{Dataset Selection}
The main analysis in this section focuses on the four representative datasets highlighted in \cref{fig:contrast} during the performance evaluation against notable frameworks.

To further test the model's generalization capabilities, the evaluation was extended to four additional external PTM datasets. This supplementary set includes methylation-G, which is severely imbalanced, and methylation-R, which is class-balanced. It also includes Ubiquitin K*, a severely imbalanced dataset representing a high-frequency modification, and Crotonylation\_K, a class-balanced dataset representing a less-studied modification. The detailed experimental results for these additional datasets are presented in the Supplementary Material.

\subsubsection{Result Analysis}
The ablation study across four representative datasets, detailed in \cref{fig:ablation_combine}, confirms that the full PDeepPP model exhibits superior comprehensive performance compared to all its ablated variants. This is evident in the ROC and PR curves, where the full model (blue and orange lines) consistently envelops the curves of other versions across nearly all tasks, demonstrating a more potent classification capability regardless of the threshold. In terms of aggregate metrics, PDeepPP achieved an average ACC of 0.8937, a BACC of 0.8786, and an MCC of 0.7598, placing it significantly ahead of any model with removed or replaced components.

\paragraph{\textbf{Dynamic Embedding Enhances Task-Specific Adaptation}}
The BaseEmbedding module, equipped with an adaptive α parameter, facilitates task-specific fine-tuning. On the Anticancer\_main task, for instance, removing this embedding component (w/o embedding) results in a decrease in the MCC score from 0.4949 to 0.4768. While this particular metric shift is modest, a consistent pattern emerges in the AUC/PR AUC curve visualizations across all datasets: the curve for the complete model consistently outperforms the variant lacking the embedding, leading to more robust classification.
\paragraph{\textbf{TIM Loss Improves Minority Class Prediction}}
The mutual information term within the loss function is crucial for preventing the model from favoring the majority class on imbalanced data. This effect is observed in the Ubiquitination task, where removing the TIM loss (w/o loss) reduces the BACC from 0.9691 to 0.9615. Likewise, on the severely skewed N6-acetyllysine dataset, substituting the TIM loss with standard cross-entropy compromises the model's capacity to recognize positive instances, causing the Sensitivity (SN) to fall from 0.9073 to 0.8917.
\paragraph{\textbf{Parallel Architecture Enables Multi-Scale Feature Extraction}}
The TransLinear module is essential for capturing global dependencies, and its removal in the Anticancer\_main task triggers a steep decline in MCC from 0.4949 to 0.3611. The PosCNN module is instrumental in recognizing local motifs; its absence on the N6-acetyllysine task severely compromises the model's ability to identify positive samples, causing the SN to plunge from 0.9073 to 0.8051. Internal mechanisms are equally critical: the removal of the attention mechanism in the Ubiquitination task lowers the MCC from 0.9442 to 0.9204. Meanwhile, lacking positional encoding on the N6-acetyllysine dataset leads to a substantial drop in MCC from 0.9008 to 0.8407, underscoring that precise spatial information plays a crucial role in identifying modification sites.

\begin{figure*}[htbp]
\includegraphics[width=1\textwidth]{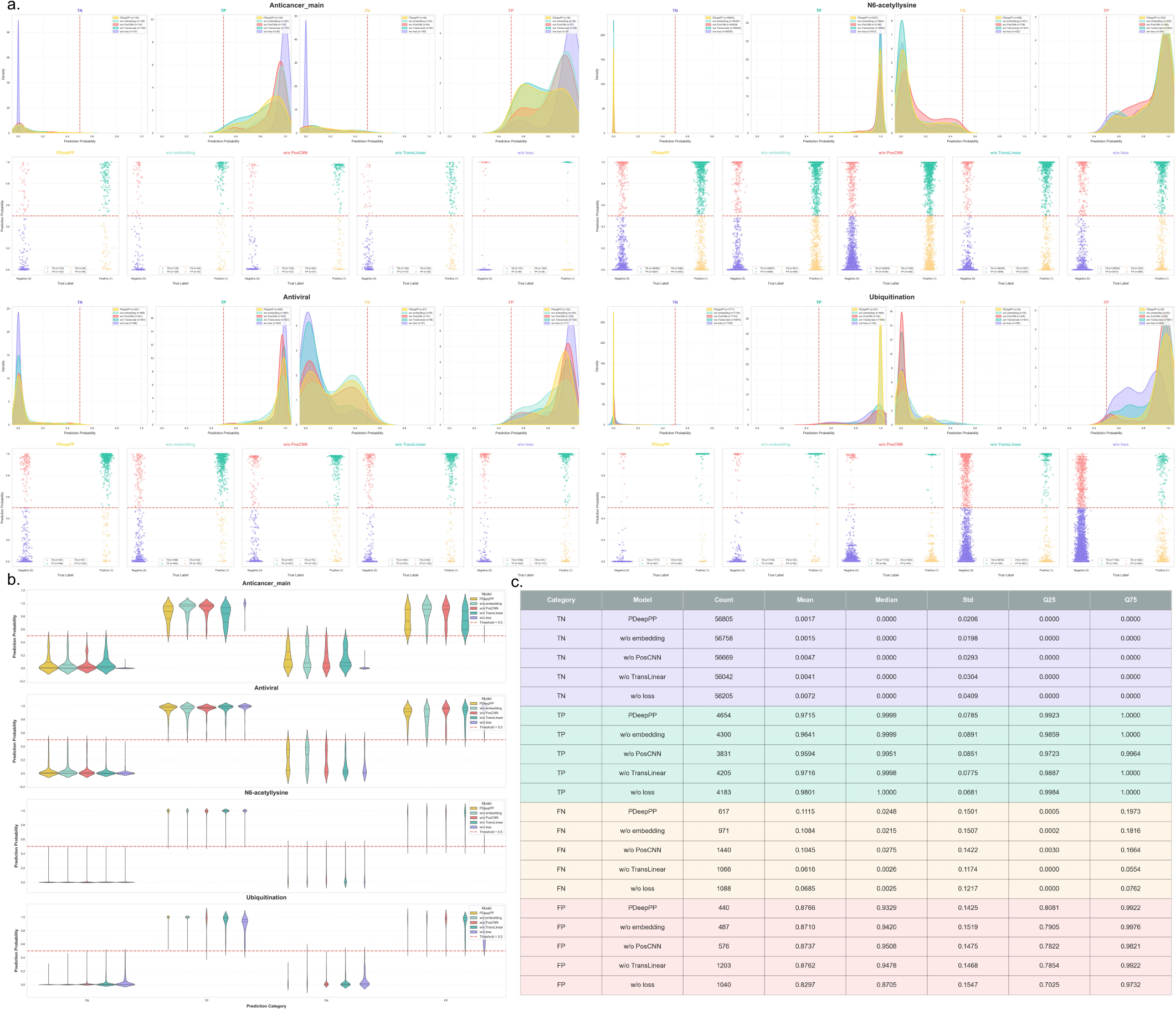} 
    \caption{Performance evaluation of prediction models across four protein datasets.
(a) Detailed visualization of prediction performance in a 2x2 layout, with density plots in the upper panel and scatter plots in the lower panel for each subplot, and the task name above the subplot. For each dataset, the top row shows kernel density estimation plots for the four prediction categories (TN, TP, FN, FP), with all five models overlaid for comparison. The bottom row presents scatter plots of true labels versus predicted probabilities for each individual model, with points color-coded by prediction category. The red dashed line represents the decision threshold at 0.5. Sample sizes for each category are indicated in the legends.(b) Violin plots comparing prediction probability distributions of the PDeepPP model and different ablation models across four datasets. Each plot displays the density distribution of prediction values for TN, TP, FN and FP. (c) Statistical summary of prediction value distribution across four prediction categories (TN, TP, FN, FP) for all models and datasets combined. The table presents the count, mean, median, standard deviation, and quartile values (Q25, Q75) for each model-category combination, with all statistical values rounded to four decimal places. The backgrounds are color-coded to correspond with the prediction categories.}
\label{fig:performance_eval}     
\end{figure*}

\subsection{Analysis of Prediction Value Distribution and Confidence for PDeepPP and its Main Variants}
To deeply evaluate the model's predictive behavior and further validate the necessity of PDeepPP's integrated design, we analyzed the distribution of prediction values output by the model on four representative datasets using the analysis presented in \cref{fig:performance_eval}. This analysis focuses on the specific performance and confidence characteristics of PDeepPP and its main ablated variants (w/o embedding, w/o PosCNN, w/o TransLinear, w/o loss) on four sample categories: True Positives (TP), True Negatives (TN), False Positives (FP), and False Negatives (FN).

The complete model achieves the highest number of correct predictions across the four datasets. Its strategy does not pursue the most extreme confidence but rather enhances overall performance through balance. PDeepPP demonstrates high confidence in its correct predictions, while the probability distribution of its incorrect predictions is closer to the decision boundary, indicating that the model is more cautious when facing uncertainty. For example, in the FP distribution for the Anticancer\_main task (\cref{fig:performance_eval}a), the probability peak of PDeepPP is further from 1.0 compared to models other than w/o TransLinear, showing a relatively lower confidence when making incorrect judgments.

The model's predictive behavior exhibits clear task dependency. According to the violin plots (\cref{fig:performance_eval}b), the TP distribution for PTM tasks is compact and highly concentrated, which corroborates the findings from the interpretability analysis—that conserved local sequence motifs provide clear signals for the model. In contrast, for BP tasks, where patterns are more diverse, the TP distribution is noticeably flatter and wider. For FN samples, which are challenging for all model variants, their probability distribution shapes do not show a consistent divergence trend among the ablated models, indicating that these misclassified samples pose a persistent challenge for all models.

The comparison with the w/o loss model directly highlights the value of the TIM loss function. The statistics table (\cref{fig:performance_eval}c) shows that the w/o loss model generated 1088 FN samples, far more than PDeepPP's 617, reflecting its tendency to predict the majority class on class-imbalanced data. Concurrently, the mean (0.0685) and median (0.0026) of the FN predictions for the w/o loss model are much lower than PDeepPP's, indicating it is more confident when incorrectly rejecting positive samples. PDeepPP also has fewer FP samples. This is because the mutual information term in the TIM loss enhances the model's feature learning for the minority class (positive samples), thereby improving the overall prediction accuracy for positive samples.

\section{Conclusion}\label{sec4}
In this study, we introduce PDeepPP, an innovative deep learning framework designed to provide a general and consistent predictive model for protein function identification. We primarily validate its effectiveness through two representative tasks: bioactive peptide identification and post-translational modification (PTM) site prediction. In the embedding stage, PDeepPP uses a dynamically adjustable parameter (α) to fuse features from the pretrained ESM-2 with task-adaptive base embeddings. Subsequently, these features are fed into a parallel feature extraction network architecture based on CNNs and Transformers. During the training phase, we adopted the TIM loss function to effectively address the prevalent issue of class imbalance in biological data.

To validate the rationality of the model design, we systematically evaluated the function of each key component. The results demonstrate that the synergy between the different parts of this hybrid architecture contributes to enhancing the final performance. Concurrently, interpretability analysis shows that PDeepPP can learn key sequence patterns with biological significance, and the features underlying its predictions are largely consistent with the actual distribution of sequence conservation, which, to some extent, confirms the model's effectiveness.

PDeepPP demonstrates excellent performance and generalization capabilities across extensive benchmark tests. In an evaluation covering 33 biological identification tasks, the model achieved results superior to existing methods on the majority of tasks. This proves that PDeepPP, as a general architecture, can maintain a high standard in peptide function prediction tasks including BPs and PTMs.

The main contribution of this research is providing a general computational platform for peptide function and PTM site prediction, reducing the need to develop specialized models for each specific task. This work can serve as a valuable baseline for developing more promising general-purpose protein sequence analysis models in the future. It is also expected to advance downstream applications, such as accelerating the discovery of novel therapeutic targets and supporting biomedical research, thereby reducing reliance on traditional experimental methods.

Although PDeepPP performs well, we recognize that it has some limitations. The current model relies solely on one-dimensional sequence information, so integrating multi-modal data, such as three-dimensional protein structures, is an important direction for future improvement. Building on this, the model could be further extended to multi-label or multi-task learning, enabling simultaneous prediction of multiple functions or modifications on the same peptide sequence. Furthermore, developing more advanced explainable AI techniques will help extract more specific and biologically verifiable sequence patterns from the model.

\section{Data and Code Availability}
The source code and preprocessed data supporting this study are openly available on GitHub at \url{https://github.com/fondress/PDeepPP} and \url{https://huggingface.co/fondress/PDeppPP}. Trained models, evaluation scripts, and implementation details are included in the repository to ensure reproducibility.

\bibliographystyle{unsrt}
\bibliography{reference}

\end{document}